\documentclass[final]{cvpr}

\usepackage[pagebackref=true,breaklinks=true,colorlinks,bookmarks=false]{hyperref}

\usepackage{times}
\usepackage{epsfig}
\usepackage{graphicx}
\usepackage{amsmath}
\usepackage{amssymb}
\usepackage{multirow}
\usepackage{subfigure}
\usepackage{bm}
\usepackage{enumitem}
\graphicspath{{Images/}}

\def\cvprPaperID{5102} 
\def\confYear{CVPR 2021}

\author{Xingyu Chen$^1$\thanks{Corresponding author, chenxingyu@kuaishou.com}\hspace{0.15in} Yufeng Liu$^{1,3}$\hspace{0.15in} Chongyang Ma$^1$\hspace{0.15in} Jianlong Chang$^2$\hspace{0.15in} Huayan Wang$^1$\\ Tian Chen$^1$\hspace{0.3in} Xiaoyan Guo$^1$\hspace{0.25in} Pengfei Wan$^1$\hspace{0.3in} Wen Zheng$^1$ \\
$^1$Y-tech, Kuaishou Technology   \hspace{0.3in}   $^2$Huawei Cloud \& AI \\  $^3$SEU-ALLEN Joint Center, Institute for Brain and Intelligence, Southeast University, China. \\ 
}

\newcommand{\nothing}[1]{}

\newcommand{\loss}{\mathcal{L}}
\newcommand{\lossmesh}{\loss_{mesh}}
\newcommand{\lossposethree}{\loss_{pose3D}}
\newcommand{\lossposetwo}{\loss_{pose2D}}
\newcommand{\losssilhouette}{\loss_{sil}}
\newcommand{\lossnormal}{\loss_{norm}}
\newcommand{\lossedge}{\loss_{edge}}
\newcommand{\losstotal}{\loss_{total}}

\begin{document}

\title{Camera-Space Hand Mesh Recovery via Semantic Aggregation \\ and Adaptive 2D-1D Registration}
\maketitle

\begin{abstract}
    Recent years have witnessed significant progress in 3D hand mesh recovery. Nevertheless, because of the intrinsic 2D-to-3D ambiguity, recovering camera-space 3D information from a single RGB image remains challenging. To tackle this problem, we divide camera-space mesh recovery into two sub-tasks, i.e., root-relative mesh recovery and root recovery. First, joint landmarks and silhouette are extracted from a single input image to provide 2D cues for the 3D tasks. In the root-relative mesh recovery task, we exploit semantic relations among joints to generate a 3D mesh from the extracted 2D cues. Such generated 3D mesh coordinates are expressed relative to a root position, i.e., wrist of the hand. In the root recovery task, the root position is registered to the camera space by aligning the generated 3D mesh back to 2D cues, thereby completing camera-space 3D mesh recovery. Our pipeline is novel in that (1) it explicitly makes use of known semantic relations among joints and (2) it exploits 1D projections of the silhouette and mesh to achieve robust registration. Extensive experiments on popular datasets such as FreiHAND, RHD, and Human3.6M demonstrate that our approach achieves state-of-the-art performance on both root-relative mesh recovery and root recovery. Our code is publicly available at \url{https://github.com/SeanChenxy/HandMesh}.
\end{abstract}

\section{Introduction}


\begin{figure}[t]
\centering
\includegraphics[width=0.98\linewidth]{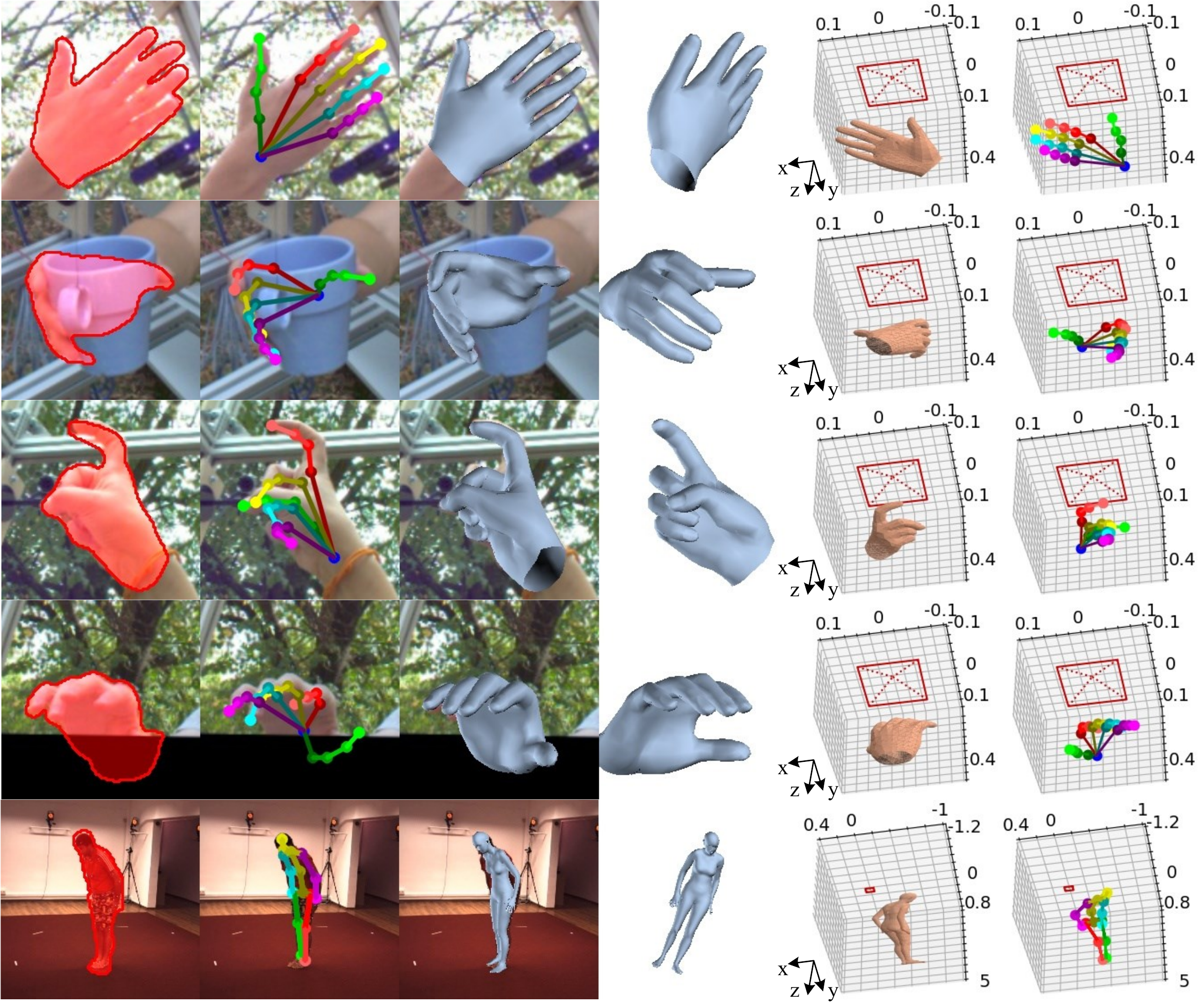}
\caption{Qualitative results of the proposed CMR. We show silhouette, 2D pose, projection of mesh, side-view mesh, camera-space mesh and pose (unit: meter). The red rectangles indicate the camera. Our method is robust enough to handle cases of occlusion, truncation, and challenging poses.}
\label{fig:vis}
\end{figure}

Monocular 3D mesh recovery has attracted tremendous 
attention due to its extensive applications in AR/VR, human–machine interaction, {\it etc}. The task is to estimate 3D locations of mesh vertices from a single RGB image. It is particularly challenging owing to highly articulated structures, 2D-to-3D ambiguity, and self-occlusion. Significant efforts have been made recently for accurate 2D-to-3D reconstruction, including \cite{bib:HMR,bib:Kanazawa19,bib:SPIN,bib:GraphCMR,bib:Kulon19,bib:I2L,pavlakos2019expressive,bib:TexturePose,bib:Pavlakos17,bib:Pavlakos18,bib:Spurr20,bib:BodyNet}, to name a few.

Most of the aforementioned methods \cite{bib:Pose2Mesh,bib:Ge19,bib:HMR,bib:SPIN,bib:GraphCMR,bib:YoutubeHand,bib:I2L,bib:Bihand,bib:Zhou17} have difficulty in predicting absolute camera-space coordinates. Instead, they define a root (\textit{i.e.}, wrist of the hand) and estimate root-relative coordinates of the 3D mesh. In this aspect, these methods cannot be applied to many high-level tasks, \textit{e.g.},~hand-object interaction, that requires camera-space mesh information. To this end, we propose to jointly solve root-relative mesh recovery and root recovery by integrating these two sub-tasks into a unified framework, thereby bridging the gap between root-relative predictions and camera-space estimation.





RGB images consist of 2D patterns that are indirect cues of the underlying 3D structure. Therefore, 2D cues have long been leveraged to assist 3D tasks. For example, 2D pose and silhouette have been used to facilitate 3D pose regression \cite{bib:Ge19,bib:SPIN,bib:I2L,bib:Pavlakos18,bib:Zhang19,bib:Zhou17,bib:Zhou20}.

However, the relationship between 2D cues and 3D structure remains unclear. We observe that
2D joint landmarks together with their semantic relations describe the 2D pose, while the silhouette indicates the holistic 3D-to-2D projection of the hand. They have different 2D properties and should be treated in different manners in the 3D task. Inspired by these observations,  we set to explore the following aspects of the 2D-to-3D task: (1) different roles of 2D cues, (2) the reason for their different effects, and (3) how to construct more effective 2D cues.



In this paper, we propose a camera-space mesh recovery (CMR) framework to integrate the tasks of 3D hand mesh and root recovery into a unified framework. CMR consists of three phases, \textit{i.e.},~2D cue extraction, 3D mesh recovery, and global mesh registration. For 2D cue extraction, we predict joint landmarks and silhouette from a single RGB image. For mesh recovery, we introduce an Inception Spiral Module for robust 3D decoding. Moreover, an aggregation method is designed for composing more effective 2D cues. Specifically, instead of implicitly learning the relations among joints, we exploit their known relations by aggregating landmark heatmaps in groups, which proves to be effective for the subsequent 3D task. Finally, camera-space root location is obtained by a global mesh registration step that aligns the generated 3D mesh with the extracted 2D landmarks and silhouette. This step is carried out via an adaptive 2D-1D registration method that achieves robustness by leveraging matching objectives in different dimensions. Our full pipeline surpasses state-of-the-art methods in the 3D mesh and root recovery tasks.
While our approach is mainly described for hand mesh, it can be readily applied to full-body mesh as shown in the experiments.
Figure~\ref{fig:vis} demonstrates several example results of our CMR for camera-space mesh recovery.

Our main contributions are summarized as follows:
\begin{itemize}[leftmargin=*,topsep=3pt]
    \setlength\itemsep{-0.1em}
    \item We propose a novel aggregation method to collect effective 2D cues and exploit high-level semantic relations for root-relative mesh recovery.
    \item We design an adaptive 2D-1D registration method to sufficiently leverage both joint landmarks and silhouette in different dimensions for robust root recovery.
    \item We present a unified pipeline CMR for camera-space mesh recovery and demonstrate state-of-the-art performance on both mesh and root recovery tasks via extensive experiments on FreiHAND, RHD, and Human3.6M.
\end{itemize}

\section{Related Work}
\paragraph{Root-relative mesh/pose recovery.}
According to different output property, we categorize methods for single-view RGB-based mesh recovery into three types, {\it i.e.,} \emph{RGB$\rightarrow$MANO/SMPL} \cite{bib:HMR,bib:Bihand,bib:Zhang19,bib:Zhou20}, \emph{RGB$\rightarrow$Voxel} \cite{iqbal2018hand,bib:I2L,bib:Pavlakos17,bib:BodyNet}, and \emph{RGB$\rightarrow$Coord} (coordinate) \cite{bib:HopeNet,bib:Ge19,bib:GraphCMR,bib:YoutubeHand}.

MANO \cite{romero2017embodied} and SMPL \cite{loper2015smpl} are parameterized 3D models of hand and human body, factorizing 3D human mesh into coefficients of shape and pose. Tremendous literature attempts to predict these coefficients for human/hand mesh recovery. For example, Zhou \etal~\cite{bib:Zhou20} estimated MANO coefficients based on the kinematic chain and developed an inverse kinematics network to improve prediction accuracy on pose coefficients. MANO/SMPL can reconstruct 3D mesh, but they embed 3D information into a parametrized space ({\it e.g.,} PCA space), where the 3D structure is less straightforward (compared to 3D vertices).

Voxel is one type of Euclidean 3D representation, to which the canonical convolutional operator can be directly applied~\cite{bib:V2V}. Thereby, the mesh recovery task can be explored in voxels. For example, Moon \etal~\cite{bib:I2L} proposed an I2L-MeshNet by dividing voxels into three lixel spaces, where a 2.5D representation is leveraged for human mesh. The voxel/2.5D-based paradigm has impressive performance in terms of human mesh recovery because the merits of Euclidean space are fully leveraged. However, voxel/2.5D representations are not efficient enough in capturing 3D details (compared to 3D vertices).

Defferrard \etal~\cite{defferrard2016convolutional} proposed a graph convolution network (GCN) based on spectral filtering to process 3D vertices in the non-Euclidean space. Based on GCN, Kolotouros \etal~\cite{bib:GraphCMR} developed a graph convolutional mesh regressor to directly estimate the 3D coordinates of mesh vertices. Ge \etal~ \cite{bib:Ge19} also developed a graph-based method for hand mesh recovery by learning from mixed real and synthetic data. Instead of spectral filtering, Lim \etal~\cite{lim2018simple} proposed spiral convolution (SpiralConv) to process mesh data in the spatial domain. Based on SpiralConv, Kulon \etal~\cite{bib:YoutubeHand} developed an encoder-decoder structure for efficient hand mesh recovery.
We follow the \emph{RGB$\rightarrow$Coord} paradigm and explore versatile aggregation of 2D cues.

\vspace{-0.4cm}
\paragraph{Root recovery.}
In analogy to root recovery, estimation of external camera parameters has been widely studied \cite{bib:Boukhayma19,bib:HMR,bib:Zhang19}. For instance, Zhang \etal~\cite{bib:Zhang19} designed an iterative regression method to simultaneously estimate external camera parameters and MANO coefficients. However, camera parameter estimation from RGB data is an ill-posed problem, leading to relatively low generalization performance. Moon \etal~\cite{bib:RootNet} proposed RootNet to predict the absolute 3D human root. RootNet essentially modeled object size in images, but the pixel-level object area has a relatively low correlation with 3D root. Rogez \etal~\cite{rogez2017lcr} predicted both 2D and 3D pose so that the 3D root location can be obtained by aligning predicted 2D pose with projected 3D pose. We argue that this 2D-3D alignment cannot sufficiently leverage 2D information and propose an adaptive 2D-1D registration method for root recovery.

\vspace{-0.4cm}
\paragraph{2D cues in 3D shape/pose recovery.}
Researchers have long exploited 2D cues in recovering 3D human shape and body parts. Pavlakos~\textit{et~al.}~\cite{bib:Pavlakos18} utilized 2D pose to regress pose coefficients and used silhouette to estimate shape coefficients. Varol~\textit{et~al.}~\cite{bib:BodyNet} first predicted 2D joint landmarks and body part segmentation, both of which were then combined to predict 3D pose.
In this work, we aim to investigate how 2D cues work in 3D tasks and to leverage them effectively for hand mesh and root recovery.


\section{Our Method}

\begin{figure}[t]
\centering
\includegraphics[width=8cm]{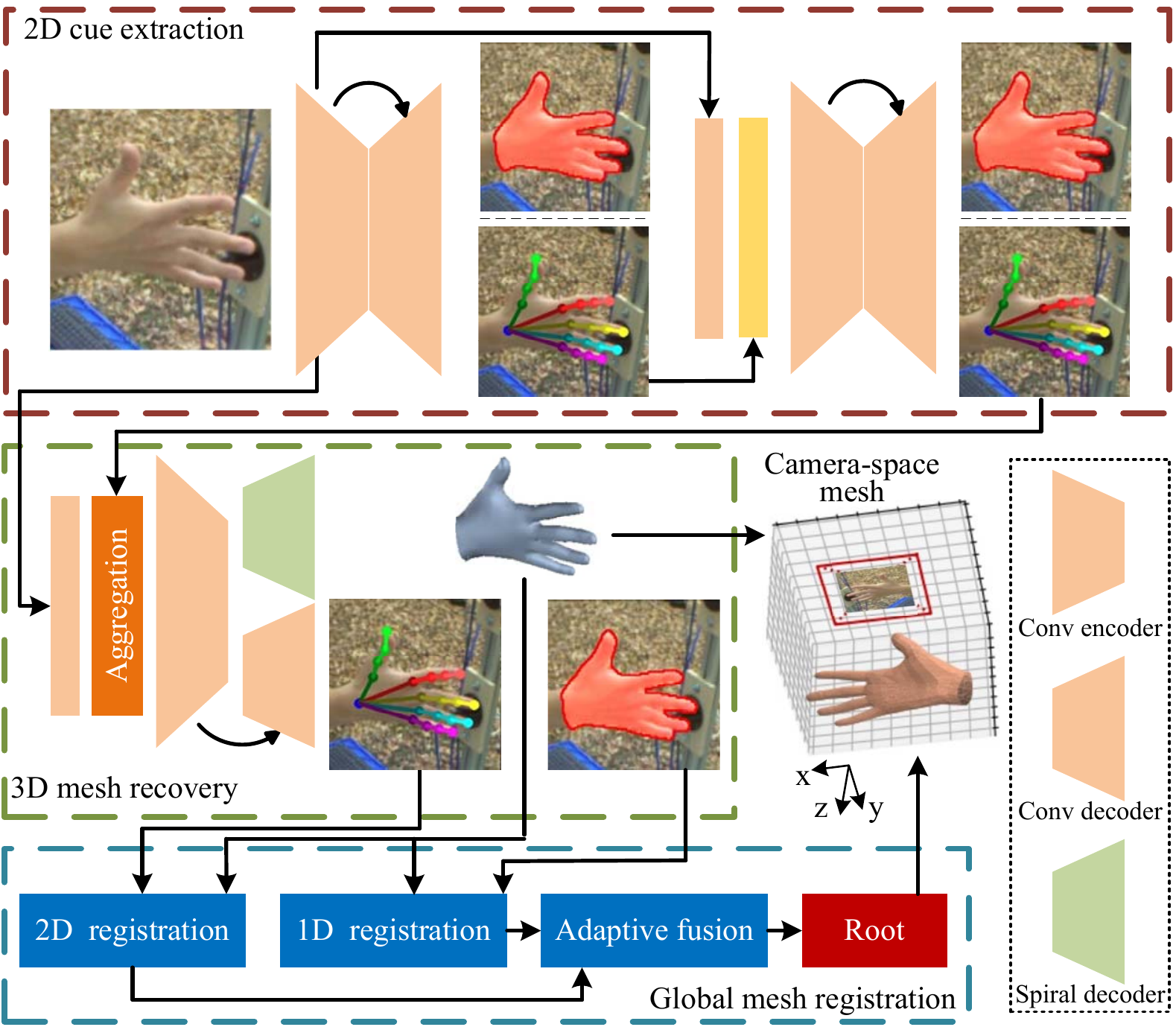}
\caption{Overview of our CMR framework.}
\label{fig:arch}
\vspace{-0.5cm}
\end{figure}

To represent a 3D mesh in camera-space, we divide it into the root-relative mesh and the camera-space root location. As shown in Figure~\ref{fig:arch}, CMR includes three phases, \textit{i.e.}, 2D cue extraction, 3D mesh recovery, and global mesh registration. In the step of 2D cue extraction, we predict 2D pose and silhouette, which are used later for both mesh and root recovery. In the step of 3D mesh recovery, we generate a root-relative mesh which is then registered to the camera space in the final phase of the pipeline.



\subsection{Mesh Recovery by Semantic Aggregation}
\paragraph{2D cues for 3D mesh recovery.}
The first phase of our pipeline extracts 2D pose and silhouette. Both 2D pose and silhouette are represented by heatmaps. To refine the 2D cues gradually, we use a multi-stack hourglass network \cite{newell2016stacked}.

The 3D mesh is defined by its {\it shape} and {\it pose} \cite{romero2017embodied}. The silhouette is a holistic 3D-to-2D projection so it captures important shape cues. However, it can hardly describe the pose accurately. On the other hand, joint landmark locations are very informative for the pose. Given their different roles, how to better combine them becomes an interesting question. Therefore, we have an insight that to improve accuracy of the subsequent 3D tasks, it is essential not only to improve the accuracy of the 2D tasks respectively, but also to better aggregate them according to their semantic relations.
Specifically, we propose to aggregate a series of 2D cues denoted as follows:

\begin{figure}[t]
\centering
\includegraphics[width=8cm]{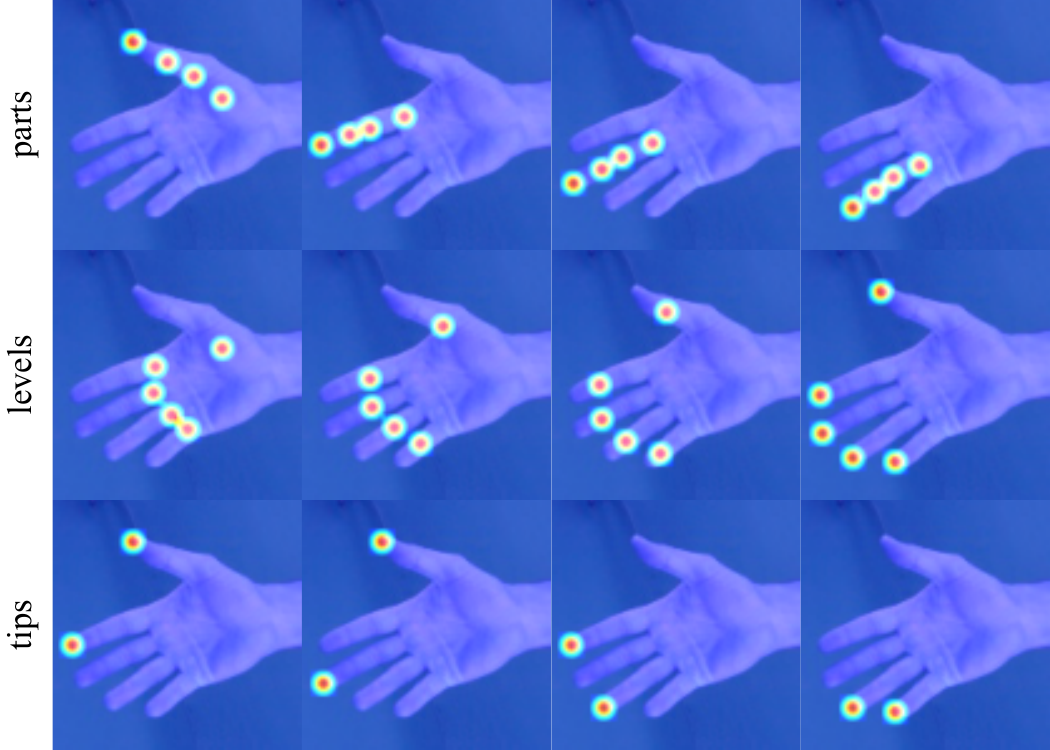}
\caption{Examples of sub-poses for high-level semantic relations. From top to bottom: part, level, and tip grouping.}
\label{fig:group}
\vspace{-0.3cm}
\end{figure}

\begin{itemize}[leftmargin=*,topsep=3pt]
    \setlength\itemsep{-0.3em}
    \item $\mathbf H_p$: $N$ heatmaps of 2D poses. Each heatmap corresponds to a joint landmark.
    \item $\mathbf H_s$: a single heatmap of the silhouette.
    \item \emph{cat}($\mathbf H_p, \mathbf H_s$): concatenating heatmaps of $\mathbf H_p$ and $\mathbf H_s$ to aggregate 2D pose and silhouette.
    \item \emph{sum}($\mathbf H_p$): combining all the joint landmarks as a single heatmap to aggregate joint locations.
    \item \emph{group}($\mathbf H_p$): concatenating $\mathbf H_p$ and tip-, part- , or level-grouped landmarks to aggregate joint semantics for high-level semantic relations.
\end{itemize}

2D silhouette and joint landmarks represent pixel-level locations in different aspects. A straightforward way of combining them is \emph{cat}($\mathbf H_p, \mathbf H_s$). Another simple baseline, \emph{sum}($\mathbf H_p$), would discard semantics of individual joints by encoding their locations in a single heatmap. Thus, comparing
$\mathbf H_p$ and \emph{sum}($\mathbf H_p$) would reveal the effect of joint semantics. These simple baselines, as we will show, are inferior to more semantically meaningful ways of aggregation: \emph{group}($\mathbf H_p$). That is, we sum the joint heatmaps in groups. As shown in Figure~\ref{fig:group}, three ways of grouping are introduced, {\it i.e.,}~by part, by level, and by tip grouping. Part grouping integrates joint landmarks on a finger, leg, arm, or torso, while level grouping integrates joint landmarks at the kinematic level \cite{bib:Zhou20}. Tip grouping integrates pairwise part tips. As a result, \emph{group}($\mathbf H_p$) forms sub-poses that exploit high-level semantic relation of 2D joints.

\begin{figure}[t]
\begin{center}
\includegraphics[width=7cm]{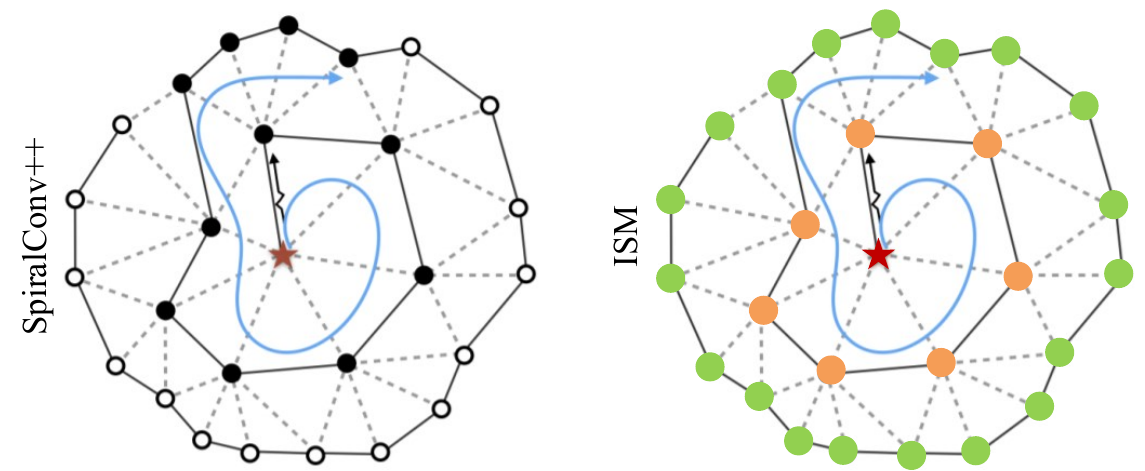}
\caption{Neighbor selection of different spiral convolutions. {\bf Left}: SprialConv++~\cite{gong2019spiralnet++}. {\bf Right}: Our method (ISM) that features hierarchical receptive fields (shown in red, orange, and green). 
}
\label{fig:spconv}
\vspace{-0.3cm}
\end{center}
\end{figure}

\vspace{-0.4cm}
\paragraph{Spiral decoder.}
The second phase of our pipeline generates the root-relative 3D mesh from the aggregated 2D cues using an improved spiral convolution decoder.

A 3D mesh $\mathcal M$ contains vertices $\mathcal V=\{\mathbf v_i=(x_i, y_i, z_i)\}_{i=1}^M$ and faces $\mathcal F$. Convolution methods for $\mathcal M$ essentially process vertex features $f(\mathbf v)$. SpiralConv \cite{lim2018simple} is a graph-based convolution operator, which processes vertex features in the spatial domain. By explicitly formulating the order of aggregating neighboring vertices, SpiralConv++~\cite{gong2019spiralnet++} presents an efficient version of SpiralConv. SpiralConv++ depends on a spiral manner of neighbor selection and adopts a fully-connected layer for feature fusion:
\vspace{-0.05cm}
\begin{equation}
\label{equ:spiral}
\begin{aligned}
    0\text{-ring}(\mathbf v) & =\{\mathbf v\} \\
    (k+1)\text{-ring}(\mathbf v) & =\mathbb N(k\text{-ring}(\mathbf v)) \setminus k\text{-disk}(\mathbf v) \\
    k\text{-disk}(\mathbf v) & =\cup_{i=0,..,k}i\text{-ring}(\mathbf v) \\
    \text{SpiralConv++}(\mathbf v) & =W(f(k\text{-disk}(\mathbf v)))+b,
\end{aligned}
\end{equation}
where $\mathbb N$ represents vertex neighborhood, $W$ and $b$ are weights and bias shared for $\forall \mathbf v\in \mathcal V$.

\begin{figure}[t]
\begin{center}
\includegraphics[width=8cm]{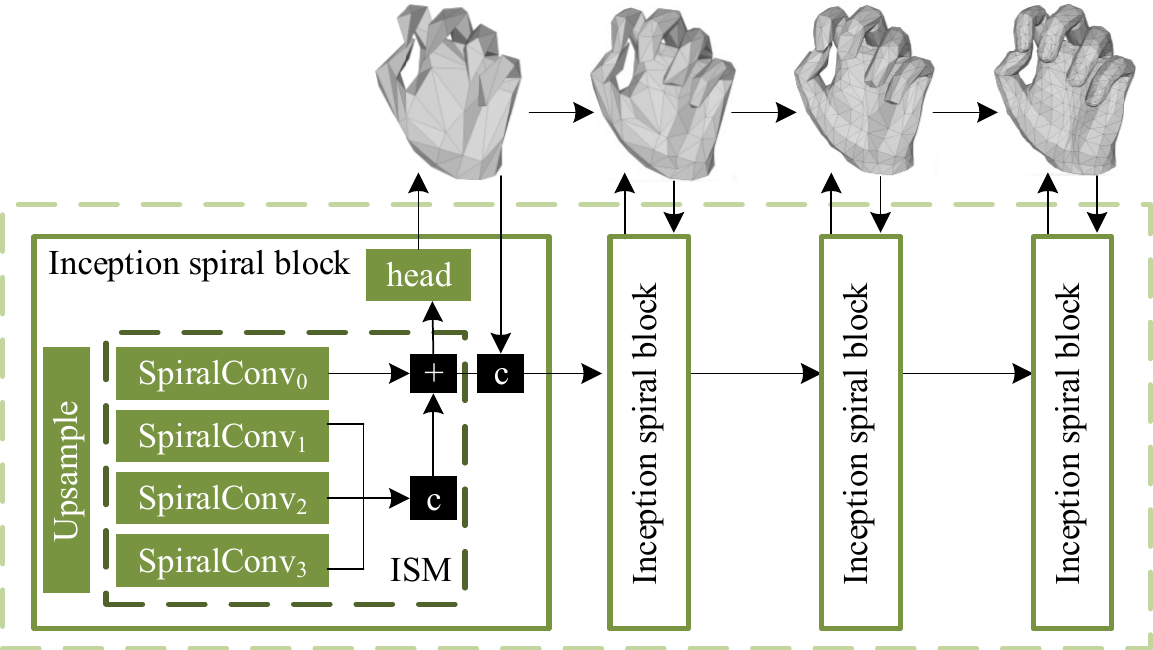}
\caption{Spiral decoder. We improve the spiral decoder by ISM, multi-scale mechanism, and self-regression.}
\label{fig:spdecoder}
\vspace{-0.6cm}
\end{center}
\end{figure}

As shown in Figure~\ref{fig:spconv}(left), SpiralConv++ collects vertex neighbors (black dots) of the cell (red star) in a spiral manner. Then these neighbors are treated indiscriminately by a fully connected layer. Inspired by Inception and residual models \cite{he2016deep,szegedy2015going}, we design an Inception Spiral Module (ISM) to enhance the receptive field of SpiralConv. Specifically, as shown in Figure~\ref{fig:spconv}(right), we distinguish neighbors according to the spiral hierarchy (red, orange, and green dots) and adopt parallel layers with diverse receptive field for 3D decoding. The ISM can be described as
\vspace{-0.05cm}
\begin{equation}
\begin{aligned}
    o_i(\mathbf v) & = W_if(i\text{-disk}(\mathbf v)) +b_i \quad i=0,1,2,3 \\
    \text{ISM}(\mathbf v) & = o_0(\mathbf v) + [o_1(\mathbf v), o_2(\mathbf v), o_3(\mathbf v)].
\end{aligned}
\end{equation}
where $[\cdot]$ denotes concatenating. In ISM we keep the number of parameters manageable by controlling the channel size of $o$.
Note that the $i\text{-disk}$ for each vertex may contain different number of elements. Similar to SpiralConv++, we truncate it to obtain a fixed-length sequence so that $W_i$ and $b_i$ can be shared for all the vertices.

The overall architecture of our 3D mesh decoder is shown in Figure~\ref{fig:spdecoder}. Our design improves the spiral decoder in three aspects: (1) we replace SpiralConv with ISM; (2) we leverage multi-scale prediction and coarse-to-fine fusion; and (3) we introduce a self-regression mechanism by concatenating scale-level predictions with the same-scale feature. Meanwhile, we use a convolutional decoder which runs in parallel with the spiral decoder to refine the estimation of 2D pose and silhouette.



\begin{figure}[t]
\begin{center}
\includegraphics[width=8cm]{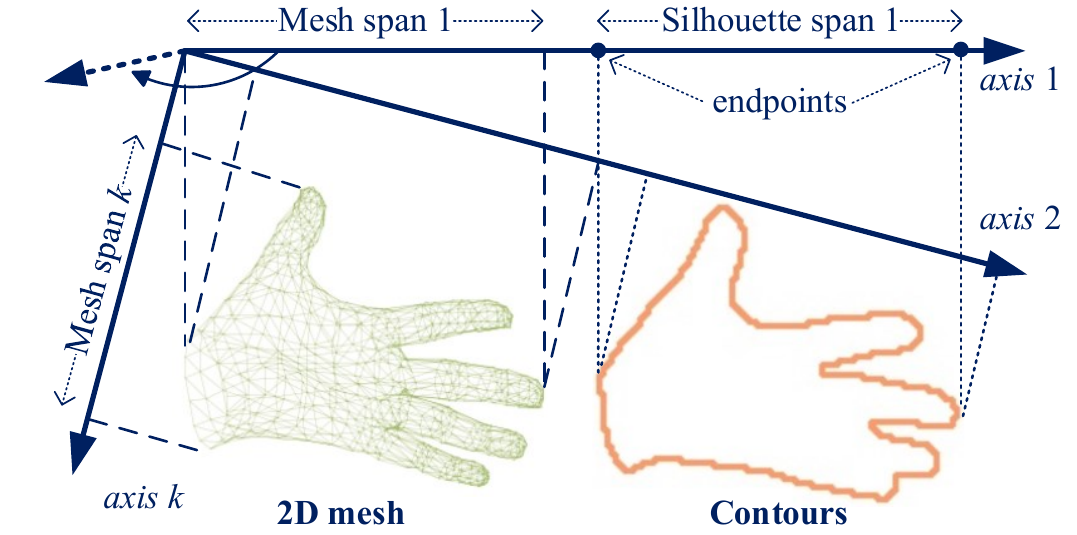}
\caption{Multiple 1D projections for the mesh and silhouette.}
\label{fig:margin}
\vspace{-0.6cm}
\end{center}
\end{figure}

\subsection{Root Recovery by Global Mesh Registration}

\paragraph{1D projections.}
The silhouette reflects holistic 3D-to-2D projection and contains strong geometric information for root recovery.
Given the intrinsic matrix $K$ of the camera, predicted 3D vertices $\mathcal V$ can be projected into the 2D space by $K\mathcal V$, resulting in a 2D mesh which consists of 2D vertices $\mathcal V^{2D}=\{\mathbf v^{2D}_i=(x^{2D}_i, y^{2D}_i)\}_{i=1}^M$ with the original connectivity. Ideally, the 2D mesh should align well with the silhouette. However, they cannot be directly aligned because (1) the large number of 2D points leads to prohibitive computational cost, \emph{i.e.}, the silhouette contains thousands of pixels while the 2D mesh has 778 (for hand) or 6,890 (for human body) vertices; (2) 2D mesh vertices have no explicit correspondence with respect to the silhouette.


To overcome these difficulties, we design a group of 1D projections to align the silhouette with the 2D mesh. First, the silhouette is converted into contours $\mathcal C=\{\mathbf c=(x^c_i, y^c_i)\}_{i=1}^C$ by edge detection \cite{canny1986computational}. We define a set of 1D axes $\mathcal A =\{\mathbf a_j=(x^a_j, y^a_j)\}_{j=1}^{A}$. These axes have unit length and are uniformly distributed, \textit{i.e.}, the angle difference between neighboring axes is $\pi/A$. As shown in Figure~\ref{fig:margin}, we project contours and 2D vertices onto $\mathbf a_j$, resulting in their 1D span, \textit{i.e.}, $\mathcal S^\mathcal C_j=\{\mathbf c_i\cdot \mathbf a_j\}_{i=1}^C$ and $\mathcal S^{\mathcal V^{2D}}_j=\{\mathbf v^{2D}_i\cdot \mathbf a_j\}_{i=1}^M$.

\begin{figure}[t]
\begin{center}
\includegraphics[width=8cm]{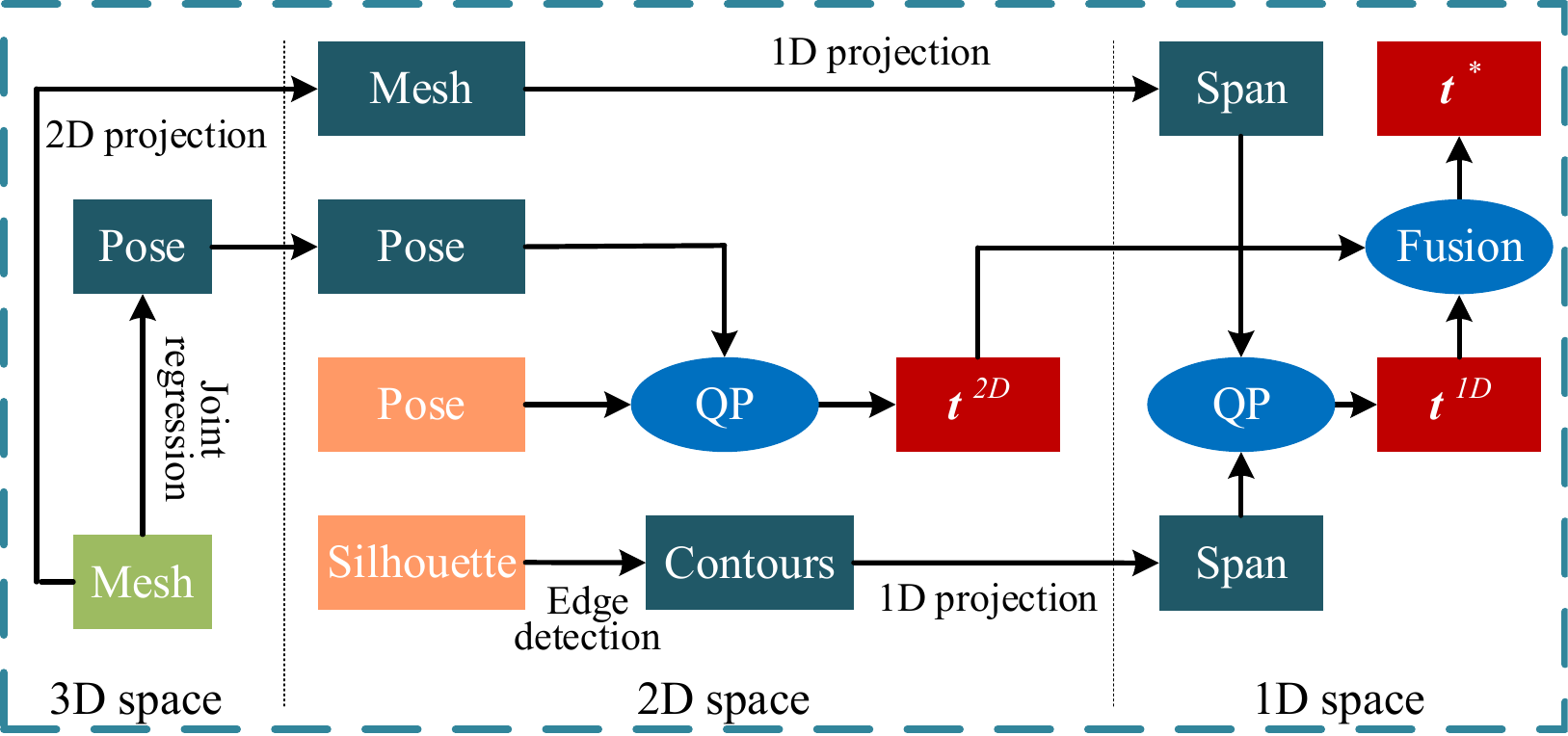}
\caption{Adaptive 2D-1D registration. For root recovery, 2D registration aligns 3D mesh with 2D pose while 1D registration aligns 3D mesh with silhouette. ``QP'' indicates quadratic programming. The green and orange elements are from the network.}
\label{fig:2d1d}
\vspace{-0.4cm}
\end{center}
\end{figure}

\paragraph{Adaptive 2D-1D registration.}
We denote the camera-space root by $\mathbf t=(x^r, y^r, z^r)$ and 2D joint landmark predictions by $\mathcal P=\{\mathbf p_i=(x^p_i,y^p_i)\}_{i=1}^N$. Given $K$ and joint regressor $J$ defined by MANO \cite{romero2017embodied} or SMPL \cite{loper2015smpl}, the camera-space 3D vertices $\mathcal V+\mathbf t$ can be converted into 2D joints by $\mathcal Q=KJ\mathcal (\mathcal V+\mathbf t)=\{\mathbf q_i=(x^q_i,y^q_i)\}_{i=1}^N$. Since $\mathcal P$ and $\mathcal Q$ have intrinsic correspondence, we define the energy function for 2D matching as
\begin{equation}
\vspace{-0.05cm}
\begin{array}{c}
        E_{2D}=\sum_{i=1}^N (\mathbf p_i-\mathbf q_i)^2,
\end{array}
\vspace{-0.05cm}
\end{equation}
whose solution is denoted as $\mathbf t^{2D}=\mathop{\arg\min}_{\mathbf t}  E_{2D}(\mathbf t)$. Further, we use $\mathcal V+\mathbf t$ to implement the aforementioned 3D-2D-1D projection. Camera-space $\mathcal S^{\mathcal V^{2D}}$ is thereby produced. Then the 1D correspondence is obtained using two endpoints of the 1D span. The energy function for alignment of 1D spans is defined as
\vspace{-0.1cm}
\begin{equation}
\begin{array}{c}
        E_{1D} = \sum_{j=1}^A\{(\max(\mathcal S^\mathcal C_j) -\max(\mathcal S^{\mathcal V^{2D}}_j))^2 \\[5pt]
          +(\min(\mathcal S^\mathcal C_j)-\min(\mathcal S^{\mathcal V^{2D}}_j))^2\}.
\end{array}
\end{equation}

In 1D space, $\mathbf t^{1D}=\mathop{\arg\min}_{\mathbf t}  E_{1D}(\mathbf t)$. Both 2D and 1D optimizations are based on quadratic programming \cite{boggs1995sequential}. With $\mathbf t^{1D}$ and $\mathbf t^{2D}$, we develop an adaptive method for the final root $\mathbf t^*$ with their distance $d = ||\mathbf t^{2D}-\mathbf t^{1D}||_2 $:
\vspace{-0.1cm}
\begin{equation}
\begin{aligned}
&\mathbf t^*=\left\{
    \begin{aligned}
        &\mathbf t^{1D} &\text{if } d>\delta_1\\
        &\frac{\delta_1-d}{\delta_1-\delta_2} \mathbf t^{2D}+\frac{d-\delta_2}{\delta_1-\delta_2} \mathbf t^{1D} &\text{else if } d>\delta_2 \\
        &\mathbf t^{2D} &\text{otherwise}
    \end{aligned}
    \right.
\end{aligned}
\end{equation}
where $\delta_1>\delta_2$, both of which are robust hyper-parameters according to 3D scale. Without wide search, we empirically use $0.06, 0.02$ for the hand and $1.0, 0.5$ for the body. This design attempts to sufficiently leverage the merits of 2D-1D registration. It is known that 2D pose is more fragile than silhouette. Hence, 2D pose is prone to be erroneous when there is a huge 2D-1D discrepancy, and silhouette is more dependable if $d>\delta_1$.
From another perspective, joint correspondence is more explicit than that of 1D projection.
Thereby, the 2D process is more reliable if 2D and 1D results are similar ($d<\delta_2$). The whole process of adaptive 2D-1D registration is presented in Figure~\ref{fig:2d1d}, from which we can see that geometrical information is sufficiently exploited from 3D to 1D spaces for root recovery.

\subsection{Loss Functions}
We use L1 norm for loss terms of 3D mesh/pose $\lossmesh,\lossposethree$, and our 2D pose/silhouette losses $\lossposetwo,\losssilhouette$ are based on binary cross entropy (BCE).
We adopt normal loss $\lossnormal$ and edge length loss $\lossedge$ for smoother reconstruction \cite{bib:Ge19}.
Formally, we have
\begin{equation}
\begin{aligned}
    & \lossmesh = ||\mathcal V-\mathcal V^\star||_1, \lossposethree = ||J\mathcal V-\mathcal J^\star||_1, \\
    & \lossposetwo = \text{BCE}(U, U^\star), \losssilhouette = \text{BCE}(S, S^\star), \\
    & \lossnormal =\sum_{\mathbf k\in \mathcal F}\sum_{(i,j)\subset \mathbf k}|\frac{\mathcal V_i-\mathcal V_j}{||\mathcal V_i-\mathcal V_j||_2}\cdot \mathbf n_\mathbf k^\star|, \\
    & \lossedge =\sum_{\mathbf k\in \mathcal F}\sum_{(i,j)\subset \mathbf k} |||\mathcal V^p_i-\mathcal V^p_j||_2 - ||\mathcal V^\star_i-\mathcal V^\star_j||_2|,
\end{aligned}
\end{equation}
where $\mathcal F, \mathcal V$ are faces and vertices of a mesh; $J,\mathcal J$ are the pose regressor and 3D joints; $\mathbf n_\mathbf k^\star$ indicates unit normal vector of face $\mathbf k$; $U,S$ are heatmaps of 2D pose and silhouette; and $\star$ denotes the ground truth. Following \cite{bib:I2L}, $U^\star$ is constructed with Gaussian distribution.

Our overall loss function is $\losstotal=\lossmesh+\lossposethree+\lambda_{p}\lossposetwo+\lambda_{s}\losssilhouette+\lambda_{n}\lossnormal+\lossedge$, where $\lambda_{p}=10,\lambda_{s}=0.5,\lambda_{n}=0.1$ are used to balance different terms.

\section{Experiments}
\label{sec:experiments}

\subsection{Experimental Setup}
We conduct experiments on several commonly-used benchmarks as listed below.
\begin{description}[leftmargin=10pt,itemsep=0pt,parsep=0pt,noitemsep,topsep=3pt]
\item[FreiHAND]~\cite{bib:FreiHAND} is a 3D hand dataset with 130,240 training images and 3,960 evaluation samples. The annotations of the evaluation set are not available, so we submit our predictions to the official server for online evaluation.
\item[Rendered Hand Pose Dataset (RHD)]~\cite{zimmermann2017learning}
consists of 41,258 and 2,728 virtually rendered samples for training and testing on hand pose estimation, respectively.
\item[Human3.6M]~\cite{ionescu2013human3} is a large-scale 3D body pose benchmark containing 3.6 million video frames with annotations of 3D joint coordinates. SMPLify-X \cite{pavlakos2019expressive} is used to obtain ground-truth SMPL coefficients. We follow existing methods~\cite{bib:Pose2Mesh,bib:HMR,bib:I2L,bib:Pavlakos17} to use subjects \emph{S1, S5, S6, S7, S8} for training and subjects \emph{S9, S11} for testing.
\item[COCO] is a wild dataset with annotations of 2D human joints. Following previous work~\cite{bib:Pose2Mesh,bib:SPIN,bib:I2L}, we use the SMPL coefficients to produce the 3D human mesh. COCO is used for training.
\end{description}

We use the following metrics in quantitative evaluations.
\begin{description}[leftmargin=10pt,itemsep=0pt,parsep=0pt,noitemsep,topsep=3pt]
\item[MPJPE/MPVPE] measures the mean per joint/vertex position error in terms of Euclidean distance (mm) between the root-relative prediction and ground-truth coordinates.
\item[PA-MPJPE/MPVPE] is the MPJPE/MPVPE based on procrustes analysis~\cite{gower1975generalized} with global variation being ignored.
\item[CS-MPJPE/MPVPE] measures MPJPE/MPVPE in the camera space for evaluation of the root recovery task.
\item[AUC] is the area under the curve of PCK (percentage of correct keypoints) vs. error thresholds.
\end{description}


\vspace{-0.5cm}
\paragraph{Implementation details.}
Our backbone is based on ResNet~\cite{he2016deep} and we use the Adam optimizer \cite{kingma2014adam} to train the network with a mini-batch size of $32$. Serving as network inputs, image patches are cropped and resized to resolutions of $\text{224}\times \text{224}$ (for the hand) or $\text{256}\times \text{256}$ (for the body). Because of different data amounts, the total number of iterations is set as 38 and 25 epochs for tasks on hand and body. The initial learning rate is $10^{-4}$, which is divided by $10$ at the 20th or 30th epoch. Data augmentation includes random box scaling/rotation, color jitter, \emph{etc.}

\subsection{Ablation Study}
\label{sec:abl}

\paragraph{Baseline.}
Our ablation studies are based on ResNet18. YoutubeHand \cite{bib:YoutubeHand} serves as the baseline, but its code and models are inaccessible. Our re-implemented model obtains $9.06$mm PA-MPVPE (see Table~\ref{tab:base}). With our spiral decoder, $8.54$mm PA-MPVPE is achieved. Thus, our designs can strengthen the robustness of 2D-to-3D decoding.

\begin{table}[t]
\small
\renewcommand{\arraystretch}{1.}
\centering
\begin{tabular}{c | c c}
\hline
Method          & PA-MPJPE $\downarrow$ & PA-MPVPE $\downarrow$  \\
\hline
YoutubeHand \cite{bib:YoutubeHand} & 8.82 & 9.06  \\
+ Our spiral decoder & 8.46 & 8.54 \\
\hline
\end{tabular}
\caption{Comparison of our spiral decoder with baseline.}
\label{tab:base}
\vspace{-0.cm}
\end{table}

\begin{table}[t]
\small
\renewcommand{\arraystretch}{1.}
\setlength\tabcolsep{4.pt}
\centering
\begin{tabular}{ c  c | c c}
\hline
    Stack 1                 & Stack 2           & PA-MPJPE $\downarrow$ & PA-MPVPE $\downarrow$  \\
\hline
$-$                                    & $-$               & 8.46     & 8.54 \\
$\mathbf H_s$                          & $-$               & 8.05     & 8.10 \\
$\mathbf H_p$                          & $-$               & 7.71     & 7.77 \\
\emph{cat}($\mathbf H_p, \mathbf H_s$) & $-$               & 7.83     & 7.95 \\
\emph{sum}($\mathbf H_p$)              & $-$               & 7.89     & 7.94 \\
\emph{group}($\mathbf H_p$)            & $-$               & 7.55     & 7.60 \\
$\mathbf H_s$                          & \emph{group}($\mathbf H_p$)  & 7.47     & 7.55 \\
$\mathbf H_p$                          & \emph{group}($\mathbf H_p$)  & 7.39     & 7.46 \\
\hline
\end{tabular}
\caption{Effects of 2D-cues aggregations after each ``Stack''.}
\label{tab:2d}
\vspace{-0.3cm}
\end{table}

\vspace{-0.5cm}
\paragraph{Effects of various 2D cues.}
This paper explores cues of 2D pose and silhouette for 3D mesh recovery, and a two-stack network is leveraged. At first, discarding the second stack, we only use one encoder-decoder structure for exposing details of 2D-cue effects. As shown in Table~\ref{tab:2d}, $\mathbf H_s$ leads to $8.10$mm PA-MPVPE\nothing{ (i.e., 8.54 vs. 8.10)} while \emph{sum}($\mathbf H_p$) reduces PA-MPVPE to $7.94$mm\nothing{ (i.e., 8.54 vs. 7.94)}.
Both $\mathbf H_s$ and \emph{sum}($\mathbf H_p$) are heatmaps that contain location information, so the locations of joint landmarks are more instructive than that of a holistic shape. This phenomenon is evident, because pose is more difficult to estimate than shape, and joint locations can directly provide cues on poses. Furthermore, $\mathbf H_p$ leads to an improved PA-MPVPE of $7.77$mm\nothing{ (\textit{i.e.}, 8.54 vs. 7.77)} by providing both joint locations and semantics. Compared to \emph{sum}($\mathbf H_p$), it is seen that joint semantics is also important. In addition, the effect of \emph{cat}($\mathbf H_p,\mathbf H_s$) is worse than that of $\mathbf H_p$\nothing{(i.e., 7.83 vs. 7.77)}. Therefore, there is no complementary benefit from 2D pose and silhouette.

\begin{figure}[t]
\centering
\includegraphics[width=0.95\linewidth]{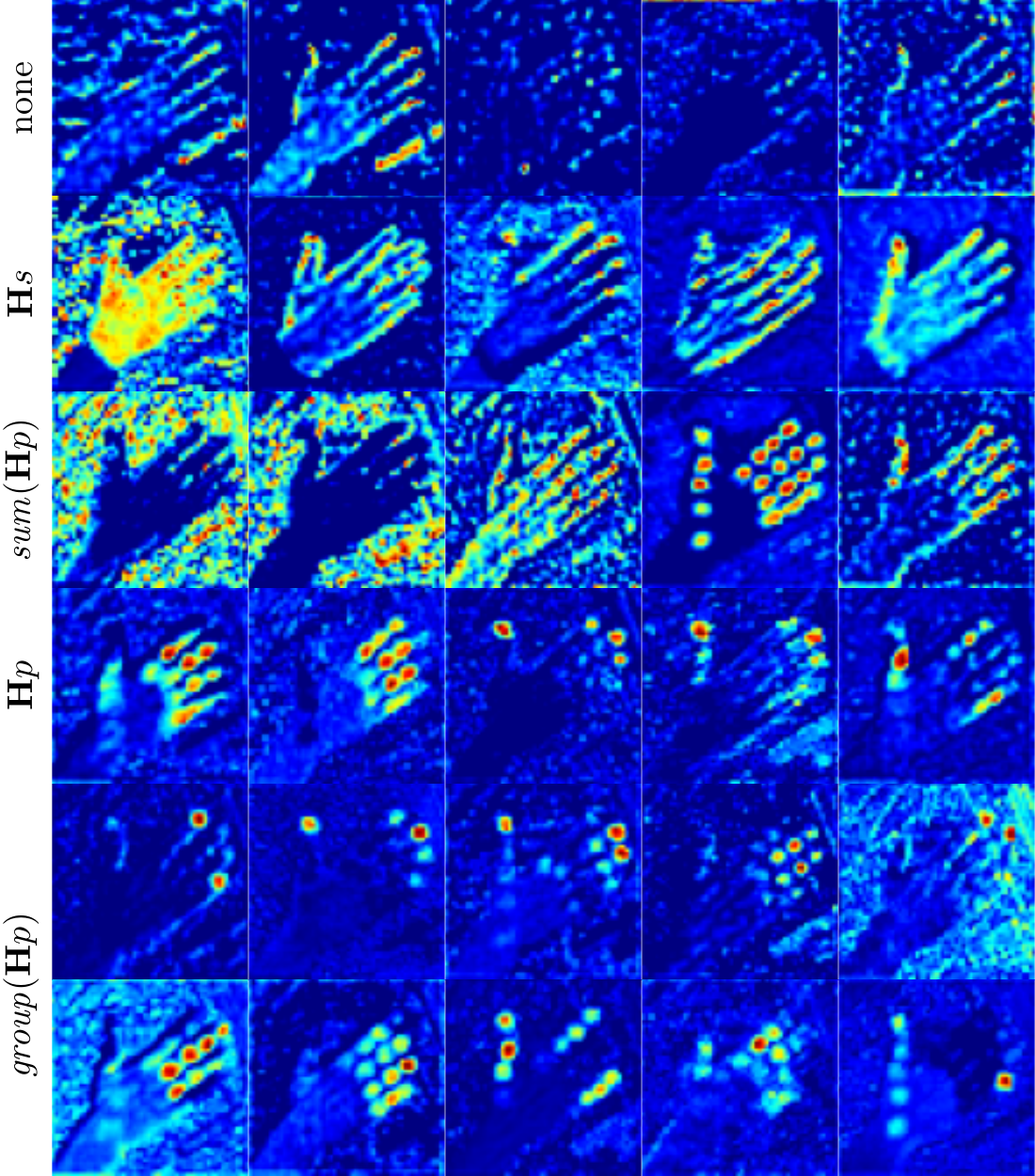}
\caption{Comparison of feature representations after various 2D cues. $\mathbf H_s$ and \emph{sum}($\mathbf H_p$) invite holistic shape and pose. $\mathbf H_p$ implies semantic relation while \emph{group}($\mathbf H_p$) induces more exhaustive joint relations by explicit semantic aggregation. The input image is shown in the first row of Figure~\ref{fig:vis}.}
\label{fig:feat}
\vspace{-0.3cm}
\end{figure}

To reveal potential reasons for different effects of 2D cues, we dissect feature representation acted by them. It is known that channel-specific feature usually embeds semantics, so we focus on typical channels in the first encoding block in the 3D mesh recovery phase.
As shown in Figure~\ref{fig:feat}, this block tends to describe trivial properties such as edges (see ``none'').
If $\mathbf H_s$ is employed, holistic 2D shapes emerge, but pose cues are ignored. With \emph{sum}($\mathbf H_p$), holistic 2D shapes still can be learned based on joint landmark locations.
Thus, this phenomenon can be the reason why there exists no complementary benefit in joint landmarks and silhouette. Moreover, holistic joint locations can also be captured with \emph{sum}($\mathbf H_p$). As for $\mathbf H_p$, although joint information is provided in separate heatmaps, the features after $\mathbf H_p$ have a tendency that multiple joints are simultaneously activated. Features shown in $\mathbf H_p$ of Figure~\ref{fig:feat} imply semantic relation of joints, which can have significant impact on 3D tasks. However, relation representation invited by $\mathbf H_p$ is not exhaustive enough. Specifically, the left two representations have similar patterns while both the 3rd and 4th representations mainly focus on the tips of thumb and middle finger.

\begin{table}[t]
\small
\renewcommand{\arraystretch}{1.}
\setlength\tabcolsep{4.2pt}
\centering
\begin{tabular}{c c c | c  c }
\hline
Parts & Levels & Tips          & PA-MPJPE $\downarrow$  & PA-MPVPE $\downarrow$ \\
\hline
            &              &            & 7.71 & 7.77  \\
\checkmark &              &            & 7.58 & 7.65  \\
            & \checkmark   &            & 7.63 & 7.69  \\
            &              & \checkmark & 7.58 & 7.65  \\
\checkmark &              & \checkmark & 7.55 & 7.60  \\
\checkmark &  \checkmark  & \checkmark & 7.62 & 7.70  \\
\hline
\end{tabular}
\caption{Effects of semantic aggregation of joint landmarks.}
\label{tab:group}
\end{table}

\begin{table}[t]
\small
\renewcommand{\arraystretch}{1.}
\centering
\begin{tabular}{c  | c | c c}
\hline
Model & Root recovery & CJ $\downarrow$ & CV $\downarrow$ \\
\hline
\multicolumn{2}{c|}{Absolute coordinate prediction}  & 77.6 & 77.4  \\
\hline
\multirow{6}{2cm}{CMR-P \\ \footnotesize (2D~AUC=0.762, mIoU=0.824)}  
                    & PnP                                    & 98.9 & 99.0 \\
                    & RootNet                                & 63.0 & 62.9  \\
                    & 2D \footnotesize{(ours)}               & 53.7 & 53.7 \\
                    & 1D \footnotesize{(ours,$A$=2)}         & 55.8 & 55.8 \\
                    & 2D+1D \footnotesize{(ours,$A$=2)}      & 52.7 & 52.7 \\
                    & 2D+1D \footnotesize{(ours,$A$=12)}     & 52.1 & 52.1 \\
\hline
\multirow{4}{2cm}{CMR-PG \\ \footnotesize (2D~AUC=\textbf{0.798}, mIoU=0.826)} 
                    & RootNet                               & 62.7 & 62.7  \\
                    & 2D \footnotesize{(ours)}              & 51.5 & 51.5 \nothing{($\downarrow$25.9)} \\
                    & 1D \footnotesize{(ours,$A$=2)}        & 54.3 & 54.3 \nothing{($\downarrow$23.1)} \\
                    & 2D+1D \footnotesize{(ours,$A$=2)}     & 51.1 & 51.1 \\
                    & 2D+1D \footnotesize{(ours,$A$=12)}    & 50.6 & 50.6 \nothing{($\downarrow$26.8)} \\
\hline
\multirow{4}{2cm}{CMR-SG \\ \footnotesize (2D~AUC=0.790, mIoU=\textbf{0.832})} 
                    & RootNet                                 & 62.8 & 62.7  \\
                    & 2D \footnotesize{(ours)}           & 52.0 & 52.0 \nothing{($\downarrow$25.4)}  \\
                    & 1D \footnotesize {(ours,$A$=2)}    & 52.6 & 52.6 \nothing{($\downarrow$24.8)} \\
                    & 2D+1D \footnotesize {(ours,$A$=2)} & 50.4 & 50.4 \\
                    & 2D+1D \footnotesize {(ours,$A$=12)}& \textbf{49.7} & \textbf{49.8} \nothing{($\downarrow$27.6)}\\
\hline
\end{tabular}
\caption{Root recovery ablation/comparison. 2D AUC and mIoU indicate the accuracy of 2D pose and silhouette. Because official evaluation do not support them, 2D AUC is about RHD while mIoU is based on our silhouette annotation. PnP \cite{ke2017efficient} and RootNet \cite{bib:RootNet} are compared. ``CJ, CV'' denote CS-MPJPE/MPVPE.}
\label{tab:abs}
\end{table}

\vspace{-0.3cm}
\paragraph{Effects of semantic aggregation.}
Instead of implied relation learning, \emph{group}($\mathbf H_p$) aims to aggregate joint semantics for explicit semantic relation. Referring to Table~\ref{tab:group}, part-, level-, and tip-based relations have instructive effects, and the integration of part- and tip-based relations leads to better performance on mesh recovery. Part-based aggregation provides within-part relations while tip-based aggregation models pairwise cross-part relations, so they are complementary for creating robust semantic relations. As a result, \emph{group}($\mathbf H_p$) surpasses $\mathbf H_p$ by $0.17$mm on PA-MPVPE\nothing{ (\textit{i.e.}, 7.60 vs. 7.77)}, {\it i.e.,} $7.60$mm. Referring to Figure~\ref{fig:feat}, compared to $\mathbf H_p$, \emph{group}($\mathbf H_p$) captures more exhaustive semantic relation, \textit{e.g.}, finger parts and various combinations of fingertips.

Besides, the two-stack network can provide more effective 2D cues, leading to better PA-MPVPE (see Table~\ref{tab:2d}). Let CMR-P, CMR-SG, and CMR-PG be models shown in the 3rd, 7th, and 8th rows of Table~\ref{tab:2d} for later analysis.

\vspace{-0.3cm}
\paragraph{Effects of adaptive 2D-1D registration.}
We train a model that directly predicts absolute camera-space coordinates for comparison. Although this operation easily suffers from overfitting, it can handle the single-dataset task, obtaining $77.4$mm CS-MPVPE (see Table~\ref{tab:abs}). Perspective-n-Point (PnP) is a standard approach that solves external parameters of a camera based on pairwise 2D-3D points~\cite{ke2017efficient}. Based on 2D joint landmarks and root-relative 3D joints from CMR-P, the PnP method can predict absolute root coordinates. However, it obtains $99.0$mm CS-MPVPE, lagging considerably behind the baseline. With our 2D or 1D registration, CMR-P achieves $53.7$mm or $55.8$mm CS-MPVPE, so both our 2D and 1D designs are valid. In detail, 2D registration is better because of unambiguous 2D-3D correspondence. Furthermore, based on our proposed adaptive
fusion, the adaptive 2D-1D registration achieves $52.7$mm CS-MPVPE. When $A=12$, the 2D-1D process can induce a better CS-MPVPE of $52.1$mm. Thus, our adaptive 2D-1D registration can sufficiently leverage 2D cues from joint landmarks and silhouette for root recovery.

\begin{table}[t]
\small
\renewcommand{\arraystretch}{1.}
\centering
\begin{tabular}{c  | c c c c}
\hline
Method       & PJ $\downarrow$ & PV $\downarrow$ & CJ $\downarrow$ & CV $\downarrow$ \\
\hline
Boukhayma \etal \cite{bib:Boukhayma19} & 35.0 & 13.2 & $-$ & $-$   \\
ObMan      \cite{hasson2019learning}   & 13.3 & 13.3 & 85.2 & 85.4 \\
MANO CNN   \cite{bib:FreiHAND}& 11.0 & 10.9 & 71.3 & 71.5   \\
YotubeHand \cite{bib:YoutubeHand}	   & 8.4  & 8.6  & $-$ & $-$   \\
Pose2Mesh  \cite{bib:Pose2Mesh}        & 7.7 & 7.8   & $-$ & $-$   \\
I2L-MeshNet\cite{bib:I2L}              & 7.4 & 7.6  & 60.3 & 60.4\\
\hline
CMR-SG (ResNet18)	    & 7.5  & 7.6 & 49.7 & 49.8 \\
CMR-PG (ResNet18)       & 7.4  & 7.5 & 50.6 & 50.6 \\
CMR-SG (ResNet50)       & 7.0  & 7.1 & \bf 48.8 & \bf 48.9 \\
CMR-PG (ResNet50)       & \bf 6.9  & \bf 7.0 & 48.9 & 49.0  \\
\hline
\end{tabular}
\caption{Results on FreiHAND. ``PJ, PV, CJ, CV'' denote PA-MPJPE, PA-MPVPE, CS-MPJPE, and CS-MPVPE, respectively.}
\label{tab:freihand}
\vspace{-0.3cm}
\end{table}

\begin{figure*}[t]
\centering
\includegraphics[width=0.95\linewidth]{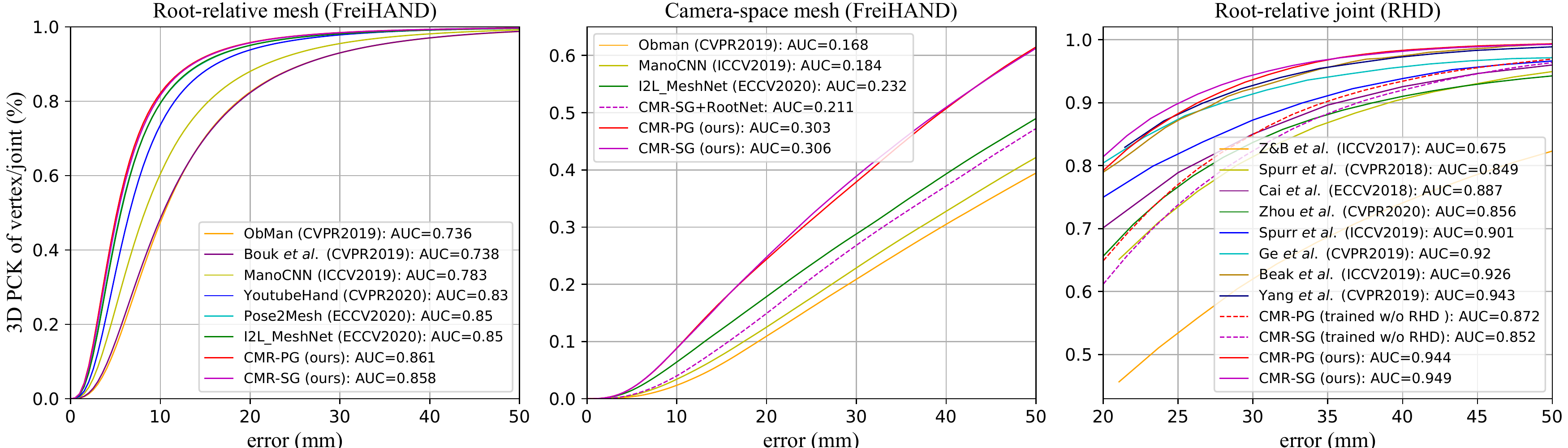}
\caption{PCK vs. error thresholds. Our CMR significantly outperforms all the other methods. 
}
\label{fig:plot}
\end{figure*}

\begin{table*}[t]
\small
\renewcommand{\arraystretch}{1.}
\setlength\tabcolsep{5.2pt}
\centering
\begin{tabular}{c  c | c c | c c | c c }
\hline
&& \multicolumn{2}{c|}{\emph{Human3.6M}} & \multicolumn{2}{c|}{\emph{Human3.6M+COCO}} \\ 
Method      & Type & MPJPE $\downarrow$ & PA-MPJPE $\downarrow$ & MPJPE $\downarrow$ & PA-MPJPE $\downarrow$ & FPS $\uparrow$ & GPU Mem. $\downarrow$\\
\hline
HMR \cite{bib:HMR} & \emph{RGB$\rightarrow$SMPL} & 184.7 & 88.4 & 153.2 & 85.5  & $-$ & $-$ \\ 
SPIN \cite{bib:SPIN}             & \emph{RGB$\rightarrow$SMPL}  & 85.6  & 55.6  & 72.9  & 51.9  & $-$ & $-$ \\ 
\hline
Pose2Mesh \cite{bib:Pose2Mesh} & \emph{Pose$\rightarrow$Coord} & \textbf{64.9}  & \underline{48.7}  & 67.9  & 49.9 & 4 & 6G \\ 
\hline
I2L-MeshNet \cite{bib:I2L}     & \emph{RGB$\rightarrow$Voxel} & 82.6 &	59.8  & \textbf{55.7}  & \textbf{41.7} & 25 & 4.6G \\ 
\hline
GraphCMR \cite{bib:GraphCMR}   & \emph{RGB$\rightarrow$Coord} & 148.0 & 104.6 & 78.3 & 59.5 & $-$ & $-$ \\ 
CMR-PG (ours, ResNet18)  & \emph{RGB$\rightarrow$Coord} & \underline{67.9} & 49.9   & 59.0  & 44.7  & \textbf{59} & \textbf{1.5G} \\ 
CMR-PG (ours, ResNet50)  & \emph{RGB$\rightarrow$Coord} & 69.8 & \textbf{47.9}  & \underline{57.3}  & \underline{42.6} & \underline{30} & \underline{1.9G} \\
\hline
\end{tabular}
\caption{Results of root-relative mesh recovery on Human3.6M.
In each column, the best number is highlighted in \textbf{bold}, and the second best number is highlighted with an \underline{underline}.
}
\label{tab:h36m}
\vspace{-0.5cm}
\end{table*}

With CMR-PG and CMR-SG, the same tendency emerges, which also validates the effectiveness of our designs. Note that CMR-PG induces better 2D pose ({\it i.e.,} $0.798$ AUC) while CMR-SG invites better silhouette ({\it i.e.,} $0.832$ mIoU), so their 2D and 1D performances are distinct.
Overall, CMR-SG achieves the best performance on root recovery and camera-space mesh recovery.

\subsection{Comparisons with Existing Methods}

We perform a comprehensive comparison on the FreiHAND dataset.
As shown in Table~\ref{tab:freihand}, our proposed CMR outperforms other methods in terms of all the aforementioned metrics.
Specifically, ResNet50-based CMR-PG achieves the best performance on root-relative mesh recovery, \textit{i.e.}, $7.0$mm PA-MPVPE and $6.9$mm PA-MPJPE.

In Table~\ref{tab:freihand}, we evaluate the root recovery performance of camera-space mesh on the FreiHAND dataset.
Our CMR outmatches ObMan \cite{hasson2019learning} and I2L-MeshNet \cite{bib:I2L} by $36.5$mm and $11.5$mm on CS-MPVPE. To the best of our knowledge, CMR achieves state-of-the-art performance on camera-space mesh recovery, \textit{i.e.}, $48.9$mm CS-MPVPE and $48.8$mm CS-MPJPE.
From Figure~\ref{fig:plot}, we can see that the proposed CMR outperforms all the compared methods on 3D PCK by a large margin.
%
With the root-relative mesh from CMR, we demonstrate that our adaptive 2D-1D registration consistently outperforms RootNet~\cite{bib:RootNet} as shown in Table~\ref{tab:abs}.

Referring to CMR-PG's PA-MPJPE and CS-MPJPE in Table~\ref{tab:freihand}, only an error of $6.9$mm is incurred by relative pose with an error larger than $40$mm from global translation, rotation, and scaling. Thus, compared with root-relative information, camera-space 3D reconstruction is more essential to improve the practicability of hand mesh recovery, and we advocate studying the camera-space problem.

In Figure~\ref{fig:plot}, we compare our CMR with several pose estimation methods \cite{baek2019pushing,cai2018weakly,bib:Ge19,spurr2018cross,yang2019aligning,bib:Zhang19,bib:Zhou20,zimmermann2017learning} on the RHD dataset. Following the criterion of PA-MPJPE, the predicted 3D pose are processed with procrustes analysis. The AUC of CMR-PG and CMR-SG is $0.944$ and $0.949$, respectively, surpassing all the other methods. In addition, we directly use the FreiHAND models for RHD test, inducing comparable AUC of $0.872$ and $0.852$. Hence, the cross-domain generalization ability of CMR is verified.

In Table~\ref{tab:h36m}, we compare our method with several state-of-the-art approaches on body mesh recovery task using the Human3.6M and COCO datasets.
Pose2Mesh~\cite{bib:Pose2Mesh} uses joint coordinates as the input and its performance downgrades slightly if the COCO dataset is added.
As an RGB-to-voxel method, I2L-MeshNet performs better when both the Human3.6M and COCO datasets are used.
By contrast, our CMR is more robust to different choices of training data and we achieve comparable or even better numbers of MPJPE and PA-MPJPE.
Moreover, our approach has faster inference speed and consumes less GPU memory.

In Figure~\ref{fig:vis}, we show several qualitative evaluation results on the FreiHAND and Human3.6M datasets.
As demonstrated in this figure, our CMR can deal with a variety of complex situations for camera-space mesh recovery, {\it e.g.,} challenging poses, object occlusion, and truncation.


\section{Conclusions and Future Work}

In this work, we aim to recover 3D hand and human mesh in camera-centered space and we present CMR to unify tasks of root-relative mesh recovery and root recovery. We first investigate 2D cues including 2D joint landmarks and silhouette for 3D tasks. Then, an aggregation method is proposed to collect effective 2D cues. Through aggregation of joint semantics, high-level semantic relations are explicitly captured, which is instructive for root-relative mesh recovery. We also explore 2D information for root recovery and design an adaptive 2D-1D registration to sufficiently leverage 2D pose and silhouette to estimate absolute camera-space information. Our CMR achieves state-of-the-art performance on both mesh and root recovery tasks when evaluated on FreiHAND, RHD, and Human3.6M datasets.

In the future, we plan to integrate 2D semantic information together with biomechanical relationship for more robust monocular 3D representation.
We are also interested in extending our CMR with a human/hand detector in a top-down manner for multi-person tasks.

{\small
\bibliographystyle{ieee_fullname}
\bibliography{ref}
}

\newpage
\section*{Supplementary Materials}

\begin{figure}[h]
\centering
\includegraphics[width=\linewidth]{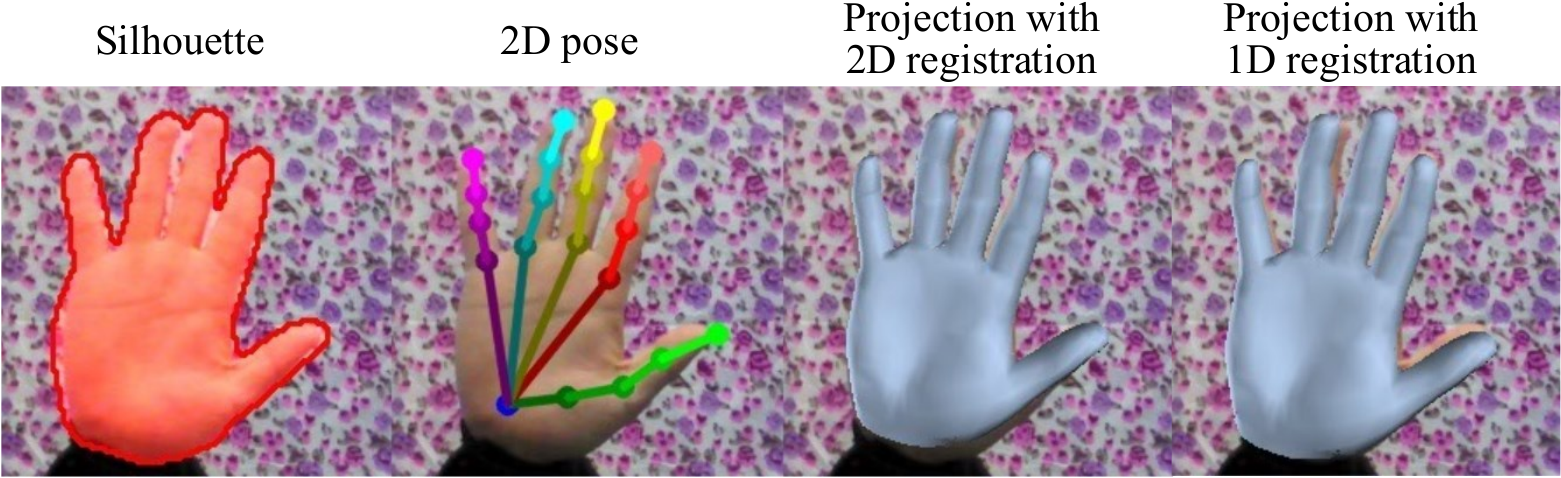}
\caption{Visualization of 2D/1D registration results.}
\label{fig:vis_2d1d}
\vspace{-0.4cm}
\end{figure}

\paragraph{Visualization of 2D/1D registration.}
As shown in Figure~\ref{fig:vis_2d1d}, the 2D registration is instructive for finger alignment while the 1D process is beneficial to holistic shape alignment. Thus, joint landmarks and silhouette have different effects on the task of root recovery, and both of them can be effectively leveraged by our adaptive 2D-1D registration scheme.

\paragraph{Effects of our designs for spiral decoder.}
We improve the spiral decoder with ISM, multi-scale mechanism, and self-regression. As shown in Table~\ref{tab:ism}, all of these three design choices are beneficial to 2D-to-3D decoding, where ISM and multi-scale mechanism has relatively more significant impact.

\begin{table}[h]
\small
\renewcommand{\arraystretch}{1.}
\setlength\tabcolsep{3.2pt}
\centering
\begin{tabular}{c c c | c  c }
\hline
ISM & Multi-scale & Self-regression     & PA-MPJPE $\downarrow$  & PA-MPJPE $\downarrow$ \\
\hline
            &              &            & 8.82 & 9.06  \\
\checkmark  &              &            & 8.73 & 8.89  \\
\checkmark  & \checkmark   &            & 8.47 & 8.56  \\
\checkmark  & \checkmark   & \checkmark & 8.46 & 8.54  \\
\hline
\end{tabular}
\caption{Effects of our designs for the spiral decoder.}
\label{tab:ism}
\vspace{-0.3cm}
\end{table}

\begin{figure}[!t]
\centering
\includegraphics[width=0.85\linewidth]{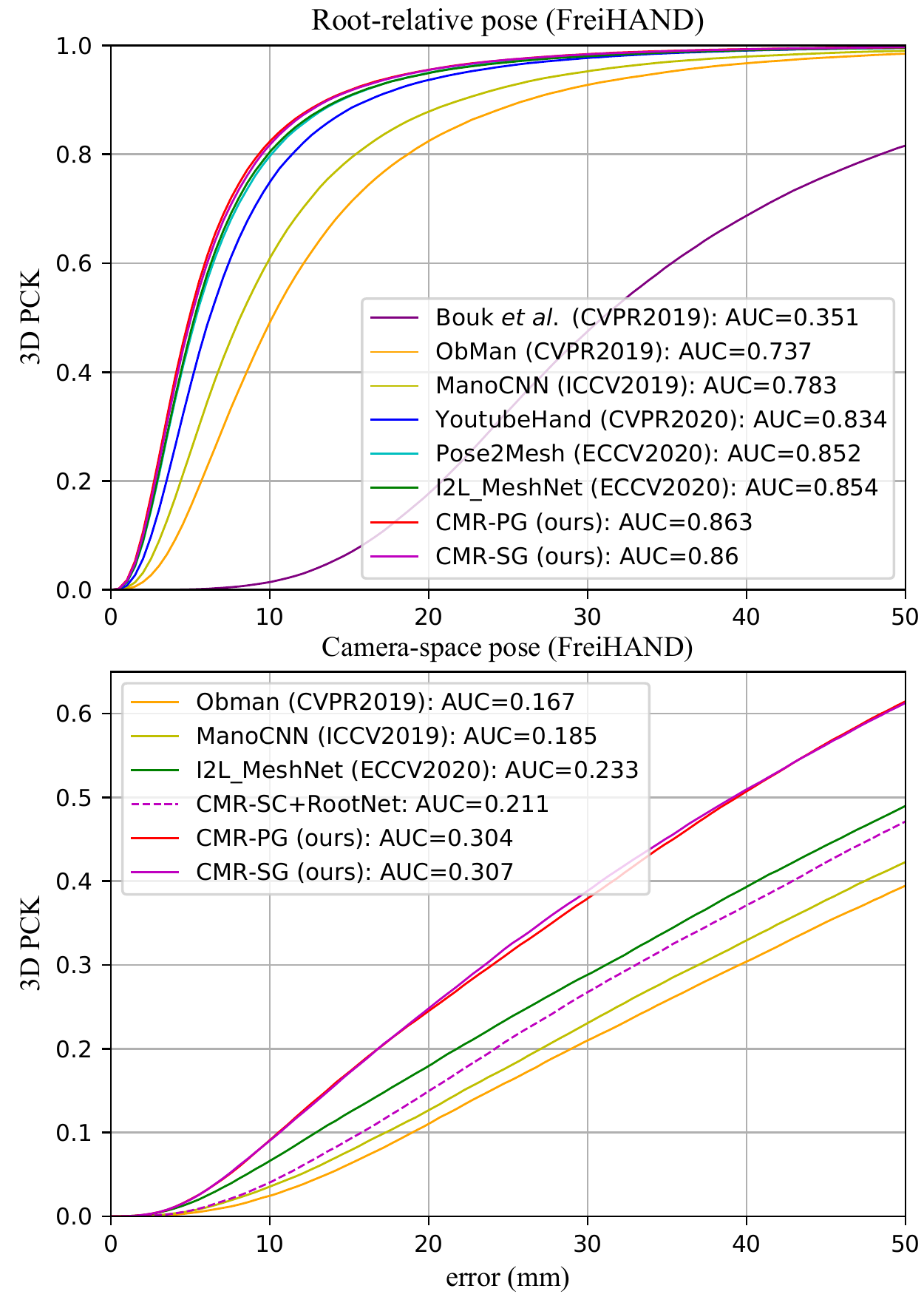}
\caption{Pose PCK vs. error thresholds. Our CMR outperforms existing methods by a large margin.}
\label{fig:freihand_pose}
\vspace{-0.4cm}
\end{figure}

\begin{table*}[!t]
\renewcommand{\arraystretch}{1.}
\small
\centering
\begin{tabular}{c  c | c c c c c c}
\hline
Method       & Backbone   & PA-MPJPE $\downarrow$ &J-AUC $\uparrow$ & PA-MPVPE $\downarrow$ & V-AUC $\uparrow$ & F@5 mm $\uparrow$ & F@15 mm $\uparrow$ \\
\hline
Boukhayma \etal (CVPR2019) \cite{bib:Boukhayma19} & ResNet50 & 35.0 & 0.351 & 13.2 & 0.738 & 0.427 & 0.895 \\
ObMan (CVPR2019) \cite{hasson2019learning}	                 & ResNet18	& 13.3 & 0.737 & 13.3 & 0.736 & 0.429 & 0.907 \\
MANO CNN (ICCV2019) \cite{bib:FreiHAND}	         & ResNet50 & 11.0 & 0.783 & 10.9 & 0.783 & 0.516 & 0.934 \\
YotubeHand (CVPR2020) \cite{bib:YoutubeHand}	     & ResNet50	& 8.4  & 0.834 & 8.6  & 0.830 & 0.614 & 0.966 \\
Pose2Mesh (ECCV2020) \cite{bib:Pose2Mesh}          & $-$      & 7.7  & 0.852 & 7.8  & 0.850 & 0.674 & 0.969 \\
I2L-MeshNet (ECCV2020) \cite{bib:I2L}              & ResNet50 & 7.4  & 0.854 & 7.6  & 0.850 & 0.681 & 0.973 \\
\hline
CMR-SG	           & ResNet18	& 7.5  & 0.851 & 7.6  & 0.850 & 0.685 & 0.971 \\
CMR-PG             & ResNet18 & 7.4  & 0.853 & 7.5  & 0.851 & 0.687 & 0.973 \\
CMR-SG             & ResNet50 & 7.0  & 0.860 & 7.1  & 0.858 & 0.706 & 0.976 \\
CMR-PG             & ResNet50 & \bf 6.9  & \bf 0.863 & \bf 7.0  & \bf 0.861 & \bf 0.715 & \bf 0.977 \\
\hline
\hline
Method       & Backbone   & CS-MPJPE $\downarrow$ &J-AUC $\uparrow$ & CS-MPVPE $\downarrow$ & V-AUC $\uparrow$ & F@5 mm $\uparrow$ & F@15 mm $\uparrow$ \\
\hline
ObMan (CVPR2019) \cite{hasson2019learning} & ResNet18   & 85.2 & 0.168 & 85.4 & 0.167 & 0.087 & 0.305 \\
ManoCNN (ICCV2019) \cite{bib:FreiHAND}          & ResNet50	& 71.3 & 0.185 & 71.5 & 0.184 & 0.102 & 0.345 \\
I2L-MeshNet (ECCV2020) \cite{bib:I2L}      & ResNet50	& 60.3 & 0.233 & 60.4 & 0.232 & 0.132 & 0.394 \\
\hline
CMR-PG             & ResNet18 & 50.6 & 0.290 & 50.6 & 0.289 & 0.155 & 0.474 \\
CMR-SG             & ResNet18 & 49.7 & 0.295 & 49.8 & 0.294 & 0.159 & 0.481 \\
CMR-PG             & ResNet50 & 48.9 & 0.304 & 49.0 & 0.303 & 0.163 & 0.488 \\
CMR-SG             & ResNet50 & \bf 48.8 & \bf0.307 & \bf48.9 & \bf0.306 & \bf0.166 & \bf0.492 \\
\hline
\end{tabular}
\caption{Results of root-relative and camera-space mesh recovery on FreiHAND. ``J-AUC, V-AUC'' denote AUC of 3D joint and mesh vertex, respectively. ``F@5 mm, F@15 mm'' is the harmonic mean between recall and precision between two meshes \emph{w.r.t.} a specific distance threshold.}
\label{tab:freihand_full}
\end{table*}

\begin{figure}[!t]
\centering
\includegraphics[width=\linewidth]{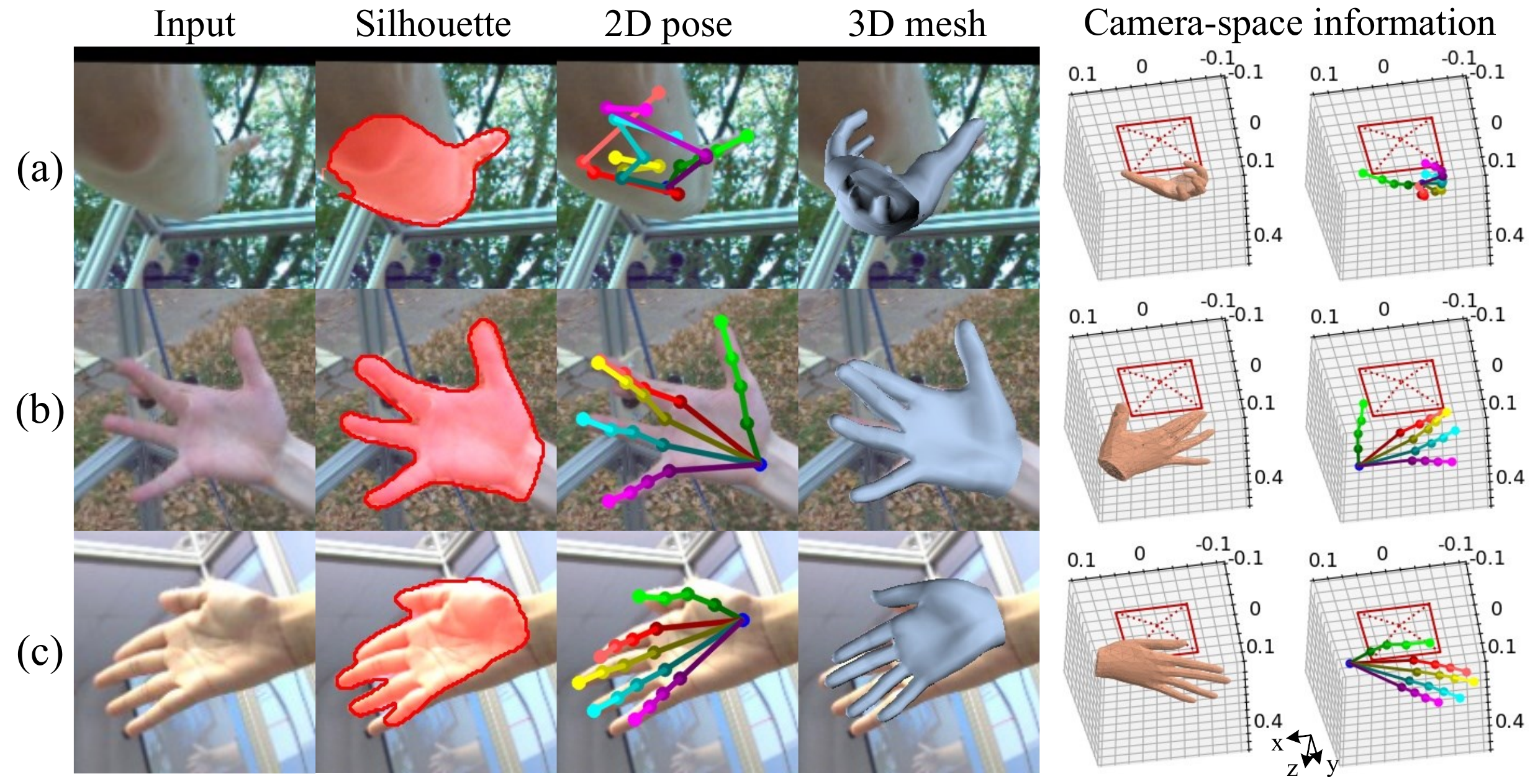}
\caption{Failure cases of CMR-SG.}
\label{fig:badcase}
\vspace{-15pt}
\end{figure}

\paragraph{Full result comparison on FreiHAND dataset.}
As shown in Table~\ref{tab:freihand_full}. For root-relative and camera-space tasks, CMR achieves state-of-the-art performance on all the merits. Figure~\ref{fig:freihand_pose} plots PCK curves of 3D joints, which can serve as a complement of Figure~\ref{fig:plot} of our main text.

\begin{figure*}[!t]
\begin{center}
\includegraphics[width=\linewidth]{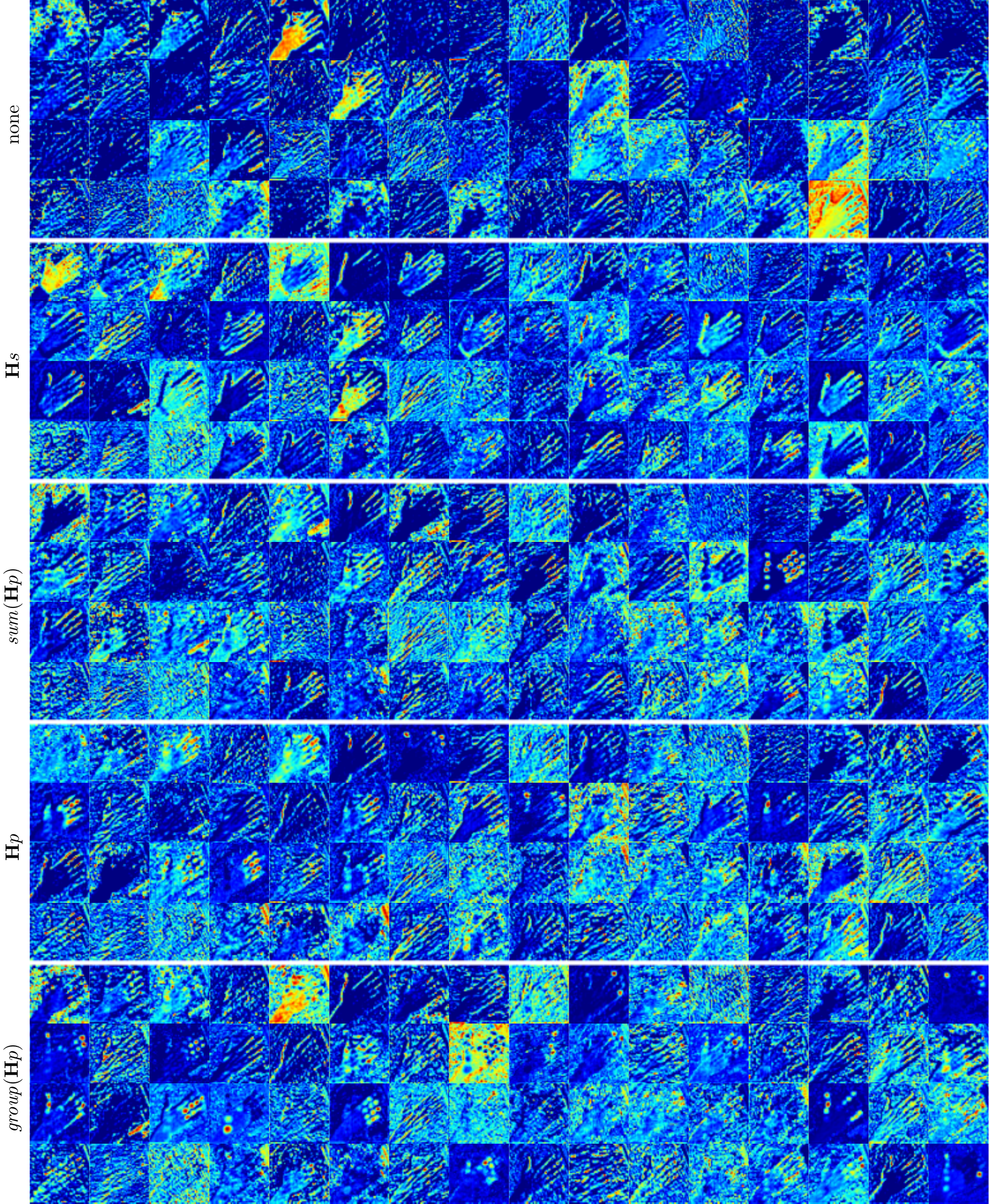}
\caption{Comparison of full feature representation after various 2D cues. \emph{group}($\mathbf H_p$) induces more semantic relation of joints. Input image is the first line of Figure~1 in the main paper.}
\label{fig:all_feat}
\end{center}
\end{figure*}

\paragraph{Full-feature representation after 2D cues.} 
Figure~\ref{fig:all_feat} serves as a complement of Figure~\ref{fig:feat} in the main text. \emph{hs} and \emph{sum(hp)} induce holistic shape and pose. In contrast, \emph{hp} invites simultaneously activated joint representation that essentially implies semantic relation. However, this relation representation is not comprehensive enough. We design \emph{group(hp)} for explicitly exploring known high-level semantics so that more comprehensive joint relations can be captured.

\begin{figure*}[!t]
\begin{center}
\includegraphics[width=16cm]{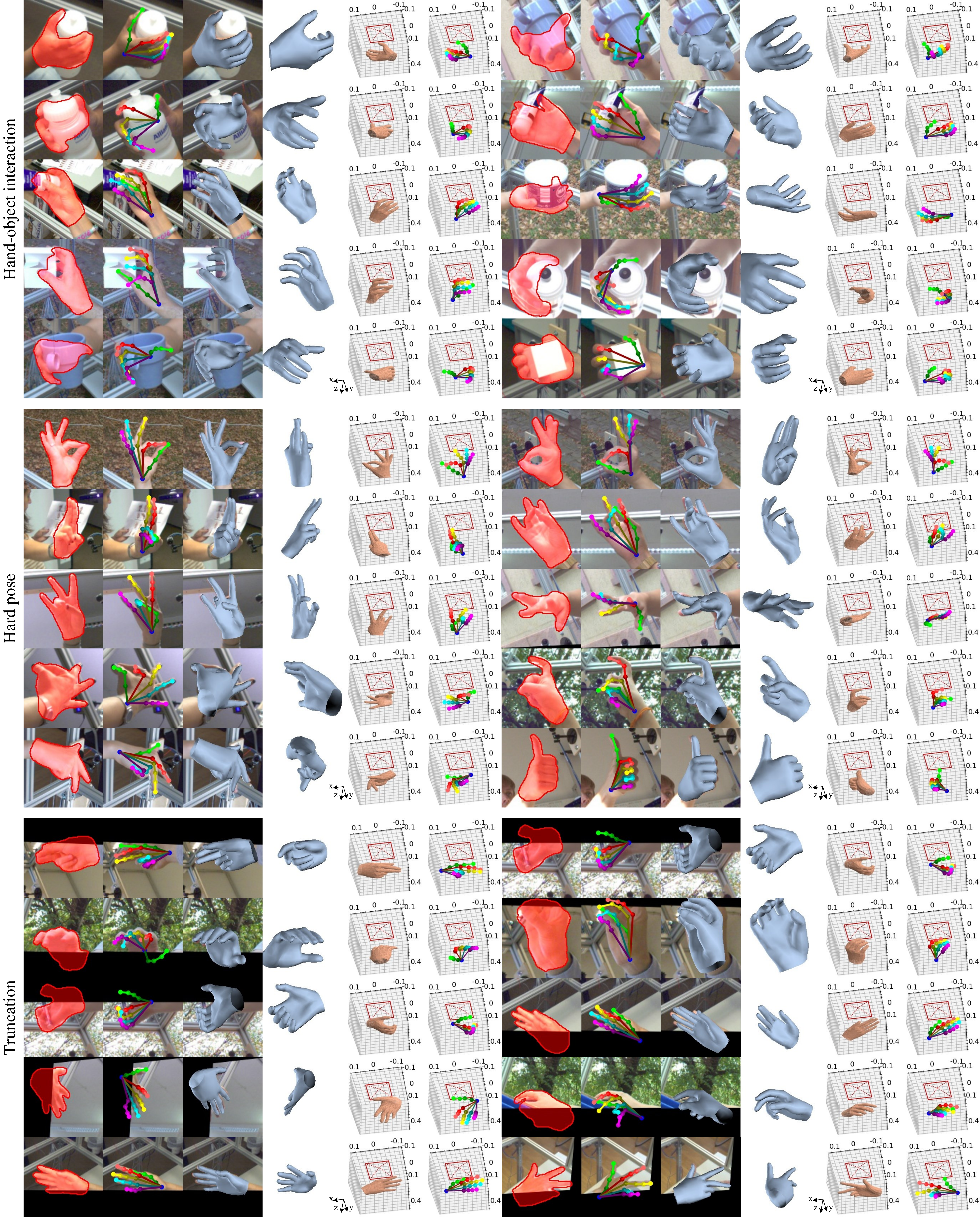}
\caption{Qualitative results on FreiHAND dataset. We can handle challenging cases of object occlusion, hard poses, and truncation.}
\label{fig:vis_freihand}
\end{center}
\end{figure*}

\begin{figure*}[!t]
\begin{center}
\includegraphics[width=16cm]{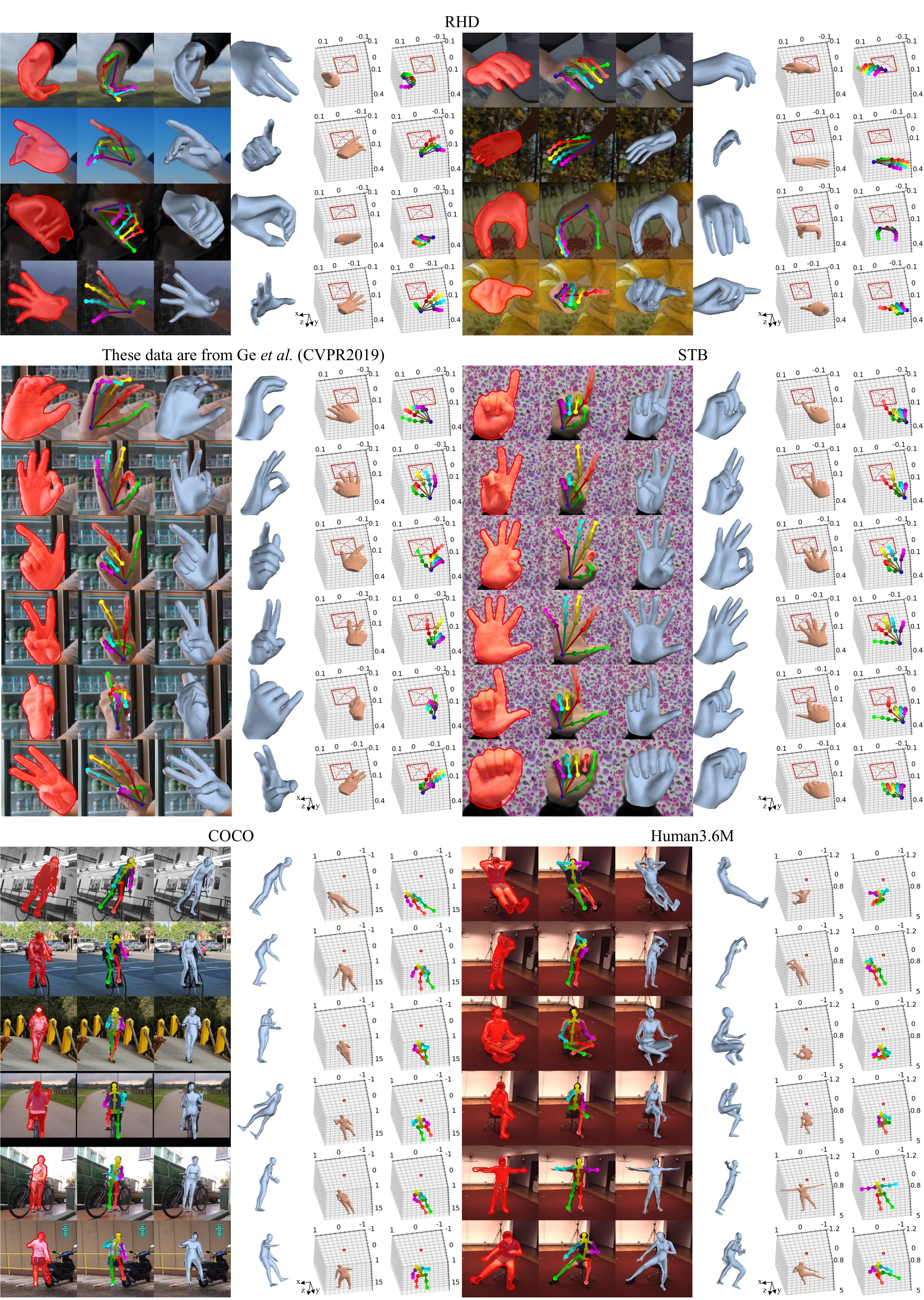}
\caption{Qualitative results on other datasets. The hand model is trained with FreiHAND only.}
\label{fig:vis_other}
\end{center}
\end{figure*}

\paragraph{More qualitative results.}
Figure~\ref{fig:vis_freihand} and \ref{fig:vis_other} illustrate comprehensive qualitative results of our predicted silhouette, 2D pose, projection of mesh, side-view mesh, camera-space mesh and pose in meter. Different from most methods, 3D roots required by image-mesh alignment are provided by CMR itself rather than ground truth, so CMR can handle images in the wild.

Referring to Figure~\ref{fig:vis_freihand}, FreiHAND's challenges include hard poses, object interactions, and truncation. Overcoming these difficulties, CMR can generate accurate silhouette, 2D pose, and camera-space 3D information.

Referring to Figure~\ref{fig:vis_other}, samples of RHD, STB, and real-world dataset released by \cite{bib:Ge19} are illustrated. We directly use the FreiHAND model for these datasets, and equally accurate predictions are obtained. Thus, CMR demonstrates superior capability of cross-domain generalization. Figure~\ref{fig:vis_other} also shows examples on Human3.6M and COCO. It can be seen that our CMR achieves reasonable results in the task of human body recovery.

\paragraph{Failure case analysis.}

Figure~\ref{fig:badcase} shows three typical failure cases of CMR-SG. 
When only a small portion of the hand is visible in the input (Figure~\ref{fig:badcase}(a)), CMR-SG predicts wrong silhouette and 2D pose. Consequently, the camera-space information is not accurate.
For cases of occlusion (\emph{e.g.}, Figure~\ref{fig:badcase}(b), in which the forefinger is completely occluded by the middle finger), although 2D pose and silhouette prediction results are still reasonable, it is difficult to obtain accurate 3D mesh since self-occlusion is challenging for the mesh recovery stage. Referring to Figure~\ref{fig:badcase}(c), strong contrast and extreme illumination change in the RGB input leads to large but consistent errors in silhouette, 2D pose, and 3D mesh prediction results.

\end{document}